\documentclass[letterpaper, 10 pt, conference]{ieeeconf}  

\IEEEoverridecommandlockouts                              

\usepackage{graphics} 
\usepackage{amsmath} 
\usepackage{amssymb}  
\usepackage{multirow}
\usepackage{multicol}
\usepackage{booktabs}
\usepackage{subfigure}
\usepackage{subcaption} 
\usepackage{svg}
\usepackage{stfloats}
\usepackage{comment}

\title{\LARGE \bf Multi-Player, Multi-Strategy Quantum Game Model for Interaction-Aware Decision-Making in Automated Driving}

\author{Karim Essalmi$^{1}$$^{,2}$, Fernando Garrido$^{1}$$^{,2}$, and Fawzi Nashashibi$^{1}$
\thanks{$^{1}$Inria Paris, 48 rue Barrault, 75013 Paris, France {\tt\small \{karim.essalmi, fernando.garrido-carpio, fawzi.nashashibi\}@inria.fr}}
\thanks{$^{2}$Valeo Mobility Tech Center, 6 rue Daniel Costantini, 94000 Créteil, France {\tt\small \{karim.essalmi, fernando.garrido\}@valeo.com}}}

\begin{document}

\maketitle
\thispagestyle{empty}
\pagestyle{empty}

\begin{abstract}
Although significant progress has been made in decision-making for automated driving, challenges remain for deployment in the real world. One challenge lies in addressing interaction-awareness. Most existing approaches oversimplify interactions between the ego vehicle and surrounding agents, and often neglect interactions among the agents themselves. A common solution is to model these interactions using classical game theory. However, its formulation assumes rational players, whereas human behavior is frequently uncertain or irrational. To address these challenges, we propose the Quantum Game Decision-Making (QGDM) model, a novel framework that combines classical game theory with quantum mechanics principles (such as superposition, entanglement, and interference) to tackle multi-player, multi-strategy decision-making problems. To the best of our knowledge, this is one of the first studies to apply quantum game theory to decision-making for automated driving. QGDM runs in real time on a standard computer, without requiring quantum hardware. We evaluate QGDM in simulation across various scenarios, including roundabouts, merging, and highways, and compare its performance with multiple baseline methods. Results show that QGDM significantly improves success rates and reduces collision rates compared to classical approaches, particularly in scenarios with high interaction. 
\end{abstract}

\section{Introduction}
Automated vehicles (AVs) are expected to provide significant benefits, including improved mobility, higher fuel efficiency, and enhanced traffic flow management \cite{makahleh2024assessing}. Despite rapid progress in the automated driving (AD) field, important challenges remain before full automation can be achieved \cite{tyagi2021autonomous}, particularly in the domain of decision-making \cite{hu2025survey}. Within the sense-plan-act paradigm, decision-making corresponds to the planning stage, where the agent must determine in real time the most suitable course of action to reach its goal. In AD, decision-making is typically divided into three levels \cite{garrido2022review, malik2022autonomous}: (a) route planning, (b) maneuver planning, and (c) trajectory planning. This study focuses on maneuver planning, which involves determining the sequence of high-level actions the ego vehicle (EV) should perform to safely reach its objectives. 

Since AVs will inevitably share the road with human drivers, algorithms must account for human behavior and adapt accordingly \cite{pan2025tgld}. To achieve this, many state-of-the-art methods model interactions mainly through safety considerations with other road users, typically employing a safety criterion to assess maneuver risk. While effective for ensuring safety, these methods suffer from two important limitations. First, they can result in overly conservative behavior \cite{fisac2019hierarchical}, potentially causing inappropriate outcomes such as the frozen robot phenomenon \cite{trautman2010unfreezing}. Second, these approaches generally neglect interactions between other agents (e.g., between two human drivers), and often assume that other agents do not adapt to the EV's behavior. Methods that explicitly address these interactions are referred to as interaction-aware approaches.

A prominent way to model interaction-awareness is through game theory, originally formalized by Von Neumann and Morgenstern \cite{von2007theory} in the economics domain and later applied to numerous domains, including biology, political science, and computer science \cite{nosheen2025game}. In the context of automated driving, game theory provides a framework to analyze the decision-making of interacting agents in conflict situations. Despite its well-established use, classical game theory (CGT) relies on strong assumptions: (a) all players are considered rational, meaning they seek to maximize their payoffs, whereas human behavior can be irrational, as discussed in \cite{song2022quantum} and \cite{lukasik2018quantum}; and (b) it fails to adapt in situations involving strategic dilemmas (e.g., when multiple Nash equilibria exist), as demonstrated in \cite{essalmiquantum}. 

To overcome these limitations, we explore quantum game theory (QGT), a subfield of classical game theory that integrates quantum mechanics concepts such as entanglement, superposition, and interference. QGT introduces new equilibrium solutions that differ from those of classical game theory \cite{wang2022advantages} and has also been shown to account for human irrational behavior \cite{zhang2021subjective, lukasik2018quantum}. However, since its introduction by Eisert et al. \cite{eisert1999quantum}, most research on QGT has remained theoretical, with few applications to real-world domains such as automated driving.
In a recent study, \cite{essalmiquantum} demonstrated the feasibility of applying QGT to decision-making in automated driving for two-player, two-action games, showing its effectiveness in resolving strategic dilemmas where classical game theory struggles. Inspired by these findings, we:
\begin{itemize}
    \item Propose a novel decision-making pipeline that integrates classical game-theoretic tools, such as strictly dominant strategies and Nash equilibria, before applying the quantum game-theoretic model. 
    \item Develop a quantum model capable of addressing multi-player, multi-strategy decision-making problems.
    \item Introduce a dynamic payoff function that adapts in real time to the evolving environment.
    \item Evaluate our approach across diverse driving scenarios (roundabouts, merging, and highways) and benchmark it against baseline methods.
\end{itemize}

The remainder of this paper is organized as follows. Section \ref{sec::relatedwork} reviews related work on decision-making in automated driving, including traditional methods, interaction-aware approaches, and prior studies in quantum game theory. Section \ref{sec::method} presents our approach. Section \ref{sec::results} reports the experimental evaluation, and Section \ref{sec::conclusion} concludes the paper. 

\par

\section{Related Work} \label{sec::relatedwork}
As we focus on the maneuver planning aspect of the decision-making, this section briefly reviews prior work that addresses maneuver planning. We also include related work on quantum game theory, which motivates our main contribution.

\subsection{Traditional Methods for Decision-Making in Automated Driving}
\textbf{Rule-based}: Rule-based techniques rely on predefined if-then statements, where specific conditions in the environment trigger corresponding actions. In \cite{aksjonov2021rule}, a rule-based decision-making algorithm was introduced for handling four-way intersections. While straightforward to implement, such approaches generally fail to generalize across diverse driving scenarios.

\textbf{Optimization}: These methods define a cost function optimized to make decisions. This function accounts for various driving parameters. For example, the study in \cite{cor-mp} determines the ego vehicle's actions by optimizing factors such as safety, passenger comfort, and efficiency. In \cite{cor-mcts}, the approach is extended to long-term planning by combining it with Monte Carlo Tree Search. Optimization provides an elegant way to address decision-making, as multiple criteria can be considered. However, the cost function must be carefully tuned for each scenario, making it difficult to generalize across diverse driving situations.

\textbf{Learning}: Since the rise of deep learning, recent research has increasingly focused on learning-based techniques \cite{talavera2021autonomous} due to their strong performance. For example, Jaritz et al. \cite{jaritz2018end} applied reinforcement learning to end-to-end autonomous racing, mapping raw images to driving actions. More recently, \cite{pang2024large} proposed a reinforcement learning approach that integrates an LLM-based driving expert to provide intelligent guidance during the training. Overall, learning-based techniques offer strong adaptability and performance but suffer from limited interpretability \cite{ghoul2023interpretable}, due to their 'black-box' nature, and often require large amounts of data to achieve reliable results.

\textbf{Probabilistic}: In probabilistic approaches, the decision-making problem is often modeled as a Partially Observable Markov Decision Process (POMDP) \cite{de2020ethical}. The main advantage of this technique lies in its ability to handle uncertainty, both aleatoric and epistemic. However, solving POMDPs in real time can be computationally demanding, and external frameworks are sometimes required. For example, the study in \cite{hubmann2019pomdp} addresses decision-making in occluded situations by modeling the problem as a POMDP and solving it with the open-source framework TAPIR \cite{klimenko2014tapir}. 

A common limitation of these traditional approaches is that they often neglect interactions with surrounding agents. Interactions are considered only indirectly, for example, through safety assessments, as in \cite{cor-mcts, cor-mp, aksjonov2021rule}, which can lead to overly conservative behavior in certain situations. To address this limitation, recent research has focused on interaction-aware methods that explicitly account for such interactions.

\subsection{Interaction-Aware Decision-Making}
As mentioned above, interaction-aware methods explicitly incorporate the interactions with surrounding agents into the decision-making process. Broadly, there are four main approaches to achieve this:

\textbf{Probabilistic}: Similar to traditional approaches, the problem can also be modeled as a POMDP in an interaction-aware manner by explicitly considering the interactions with other agents. For instance, \cite{arbabi2023decision} formulates the problem as a POMDP and solves it using Monte Carlo Tree Search. They demonstrate their approach in a merging scenario, where the ego vehicle must account for other vehicles' behaviors to safely enter the main lane.

\textbf{Cooperative}: Another approach to account for interactions is through cooperative methods. Here, the EV has access to additional information about the environment via V2X (Vehicle-to-Everything) communication. This information can include the positions and future decisions of other agents, as well as details about infrastructure states (e.g., traffic lights). For example, \cite{heshami2024towards} leverages connected vehicles and game theory, specifically the coalitional concept of game theory, to make lane change decisions. Such a technique enhances trust in the decision-making process. However, it can be costly, as it requires equipping all vehicles with V2X communication devices. 

\textbf{Learning}: Learning-based techniques are another solution to address interaction-aware decision-making. For example, \cite{chen2021midas} introduces MIDAS, a multi-agent decision-making framework based on RL for urban driving scenarios. Similarly, \cite{nan2023interaction} uses deep inverse RL to infer the reward function, which is then applied to merging scenarios.

\textbf{Game theory}: Game-theoretic approaches account for interactions by modeling agents as players, with their possible actions considered as strategies. The goal is to find an equilibrium that satisfies all players. For instance, \cite{li2023autonomous} addresses unsignalized intersections using game theory by finding the Nash Equilibrium (N.E.). Similarly, \cite{garzon2019game} handles merging scenarios using the level-k reasoning concept, where a level-k player makes decisions while assuming other agents are level-(k-1) players. Game theory is a powerful approach to interaction-aware decision-making. However, as the number of players and/or strategies increases, modeling and solving such games becomes increasingly complex.

\subsection{Quantum Game Theory}
As mentioned earlier, QGT extends classical game theory by incorporating principles of quantum mechanics. It differs from CGT in three aspects \cite{pricequantum, wiki_quantum_game_theory, morra2023quantum}: 

\begin{itemize}
    \item \textbf{Superposition}: In a classical game, a player's strategy is a discrete choice. In contrast, in a quantum game, a player's strategy can exist in a superposition of classical strategies (e.g., simultaneously \(\big|0 \rangle\) and \(\big|1 \rangle\) until a measurement is made). This means that strategies are represented as quantum states, which are linear combinations of classical strategies, thereby expanding the strategy space.

    \item \textbf{Entanglement}: In classical games, players' strategies are independent unless explicit coordination exists. In quantum games, however, players' choices can be entangled, meaning one player's choice cannot be described independently of the other's. This enables forms of coordination and correlation that do not exist in classical games. 

    \item \textbf{Quantum Moves}: In classical games, players choose among a finite set of predefined actions. In quantum games, players can apply unitary operators (or quantum gates) as their moves. This expands the set of possible strategies and might alter equilibrium outcomes compared to classical games.
    
\end{itemize}

For a more detailed comparison between CGT and QGT, the reader can refer to \cite{wang2022advantages, pricequantum}. After the brief overview of the conceptual differences between CGT and QGT, we discuss relevant studies that have applied quantum game theory.

Eisert et al. \cite{eisert1999quantum} were the first to propose a combination of game theory and quantum mechanics, introducing the Eisert-Wilkens-Lewenstein (EWL) formalism, which we adopt in our work. Applying their model to the well-known Prisoner's Dilemma, they showed that the dilemma disappears once quantum strategies are allowed. 
In \cite{meyer1999quantum}, Meyer adapted QGT to the Matching Pennies game and showed that a player adopting a quantum strategy can increase their expected payoff. They further concluded that a quantum strategy is always at least as good as a classical one. More recently, Khan et al. \cite{khan2025quantum} applied quantum game theory to the trading domain and reached a similar conclusion: a quantum player consistently holds an advantage over a classical player.
Concerning autonomous driving, only a few works have attempted to apply quantum concepts to this domain. In \cite{essalmiquantum}, the authors applied the EWL formalism to decision-making for AD. This was the first study to directly apply the EWL model to this field and to present concrete results. In particular, it was shown that in situations involving strategic dilemmas (such as merging or roundabouts), employing a quantum strategy can increase the expected payoff of a player compared to a classical strategy. 
In contrast, Song et al. \cite{song2022quantum} applied quantum concepts to decision-making for AD without using the EWL formalism. Instead, their approach was inspired by \cite{pothos2009quantum} and relied directly on the Schrödinger equation to model and solve the decision-making process.
The earlier study \cite{song2022research} focused on predicting pedestrian trajectories using quantum aspects. Specifically, it investigated whether a pedestrian would cross or not. Their results highlighted that quantum probabilities can better capture irrational behavior compared to classical probabilities. 

As seen in this subsection, only a few works have explored QGT for decision-making in automated driving. Moreover, most existing works remain at a theoretical level and lack experimental validation, motivating our study.

\section{Method} \label{sec::method}
\begin{figure*}[!t]
    \centering
    \includegraphics[width=\linewidth]{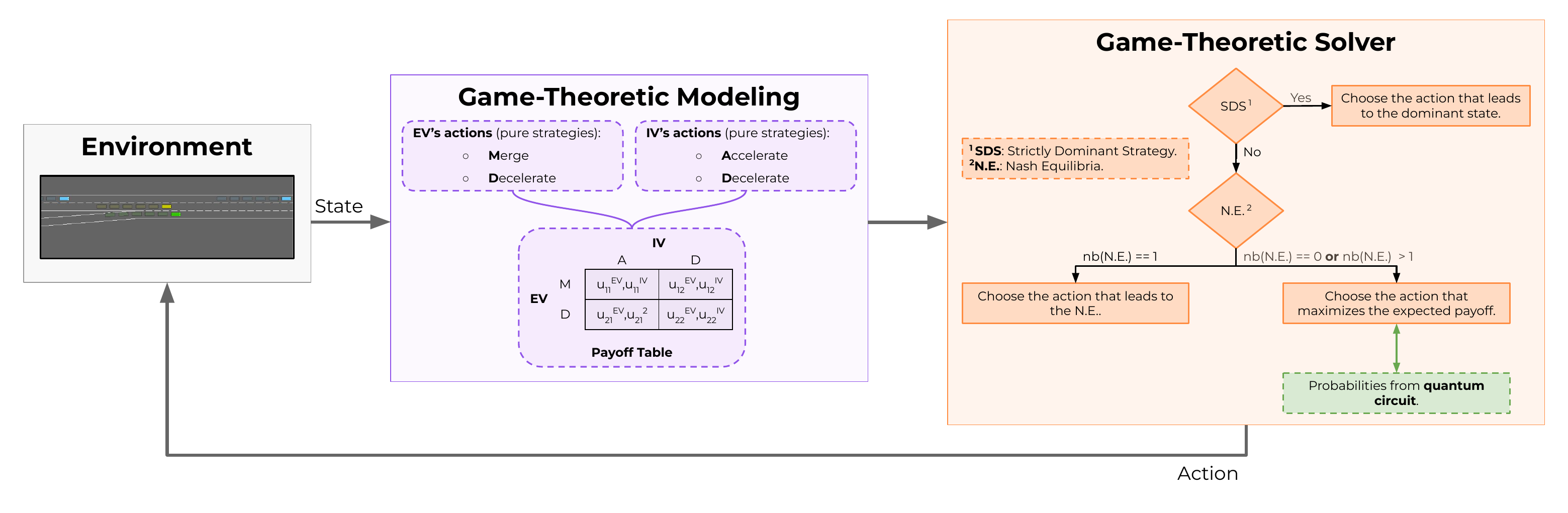}
    \caption{Pipeline of the Quantum Game Decision-Making (QGDM) framework.}
    \label{fig:pipeline}
\end{figure*}

Figure \ref{fig:pipeline} illustrates the pipeline of our framework, which is organized into three main components: environment, game modeling, and game solver. Each component plays a distinct role and is described in detail in this section. 

\subsection{Environment}
We use the highway-env simulator \cite{highway-env} as the environment in which the EV evolves. It provides the current state-space \(s \in\mathcal{S}\) in real time and passes it directly to the game-theoretic modeling component.

\subsection{Game-Theoretic Modeling}
The modeling part of our approach is defined similarly to a finite, n-person normal-form game \(\mathcal{G} = \Bigl( \mathcal{N}, \mathcal{A}, u \Bigr) \). Where:

\begin{itemize}
    \item \(\mathcal{N} = \{1, \dots, n\}\) is a finite set of players, indexed by \(i\),
    \item \(\mathcal{A}= \mathcal{A}_1 \times \dots \times \mathcal{A}_n\) is the set of all action profiles, where an action profile \(a=(a_1, \dots, a_n)\) represents the actions taken by all players, with \(a_i \in \mathcal{A}_i\) for each player \(i\),
    \item \(\mathcal{A}_i\) is the action set (i.e., the set of pure strategies) for player \(i\),
    \item \(u=(u_1, \dots, u_n)\) is the profile of utility functions for all players,
    \item \(u_i:\mathcal{A} \rightarrow \mathbb{R}\) is the utility (payoff) function for player \(i\).
\end{itemize}

Conceptually, this component represents the driving scenario as a normal-form payoff matrix. The utility function \(u_i\) for each player is adopted from the Conservation of Resources model for Maneuver Planning (COR-MP) \cite{cor-mp}, which is a utility-based maneuver planner. For each game state, a payoff value \(v \in [0,1]\) is assigned, representing how favorable the state is for a player when executing a given action, accounting for potential responses from other players. This value is computed based on multiple factors, including safety, comfort, and efficiency, with the objective of maximizing it.

\subsection{Game-Theoretic Solver}
Once the environment is modeled as a game, the next step is to solve it. As illustrated in Figure \ref{fig:pipeline}, the game is solved through the following steps:

\textbf{Step 1}: We first check whether a Strictly Dominant Strategy (SDS) exists. For a player \(i\), a strategy \(a_i^* \in \mathcal{A}_i\) is strictly dominant if, for every alternative action \(a'_i \in \mathcal{A}_i \setminus \{a_i^*\}\) and for all action profiles of the other players \(a_{-i} \in \mathcal{A}_{-i}\), the following holds:
\begin{equation}
    u_i(a_i^*,a_{-i}) > u_i(a_i',a_{-i}).
    \label{eq:dominantstrategy}
\end{equation}
In other words, an SDS is an action that yields a strictly higher payoff than any alternative, regardless of the other players' choices \cite{che2024game}. If such an action is identified, the EV selects it. Otherwise, the algorithm proceeds to Step 2.

\textbf{Step 2}: We then check if pure Nash equilibria exist in the game. A Nash equilibrium (N.E.) is an action profile in which no player can improve their payoff by unilaterally deviating from their chosen strategy, as formalized in Eq. \ref{eq:nashequilibrium} \cite{jackson2011brief}. Formally, for each player \(i\), an action profile \(a_i^* \in \mathcal{A}_i\) is considered a N.E. if: 

\begin{equation}
    u_i(a_i^*,a_{-i}^*) \ge u_i(a_i',a_{-i}^*), \quad \forall a_i' \in \mathcal{A}_i\,,
    \label{eq:nashequilibrium}
\end{equation}
where \(a_i^*\) denotes player \(i\)'s equilibrium action and \(a_{-i}^*\) represents the equilibrium actions of all other players. In some cases, multiple N.E. may exist, leading to strategic dilemmas that complicate the selection of an optimal solution \cite{essalmiquantum}. If exactly one pure N.E. is found, the EV selects the corresponding action \(a_i^*\). Otherwise, if no N.E. exists or if multiple are present, the algorithm proceeds to Step 3.

\textbf{Step 3}: As discussed by Schoemaker \cite{schoemaker1982expected}, under uncertainty, decision-makers tend to select actions that maximize their expected utility. In a game, the expected utility of a player \(i\) for a given action \(a_i\), denoted \(\mathbb{EU}_i(a_i)\) (Eq. \ref{eq::expected_utility}), is defined as the weighted sum of payoffs (or utilities), where the weights correspond to the joint probabilities of the action profiles. Specifically, \(u_i(a_i, a_{-i})\) represents the utility function, and \(p(a_i, a_{-i})\) is the joint probability that player \(i\) chooses action \(a_i\) while the other players choose \(a_{-i}\). In our framework, these probabilities are provided by the quantum circuits introduced in subsection \ref{subsec::quantummodels}.

Finally, the EV selects the action \(a_f\) that maximizes its expected utility (Eq. \ref{eq::finalaction}), thereby explicitly accounting for uncertainty through quantum probabilities.  

\begin{equation} \label{eq::expected_utility}
    \mathbb{EU}_i(a_i) = \sum_{a_{-i} \in \mathcal{A}_{-i}} p(a_i, a_{-i}) \cdot u_i(a_i, a_{-i}) \,,
\end{equation}

\begin{equation}
\label{eq::finalaction}
    a_f = \arg\max_{a_i \in \mathcal{A}_i} \mathbb{EU}_i(a_i).
\end{equation}

\subsection{Quantum Models} \label{subsec::quantummodels}

\begin{figure*}[ht]
\subfigure[QC - 2 players and 2 strategies.]{\includegraphics[width=.33\textwidth]{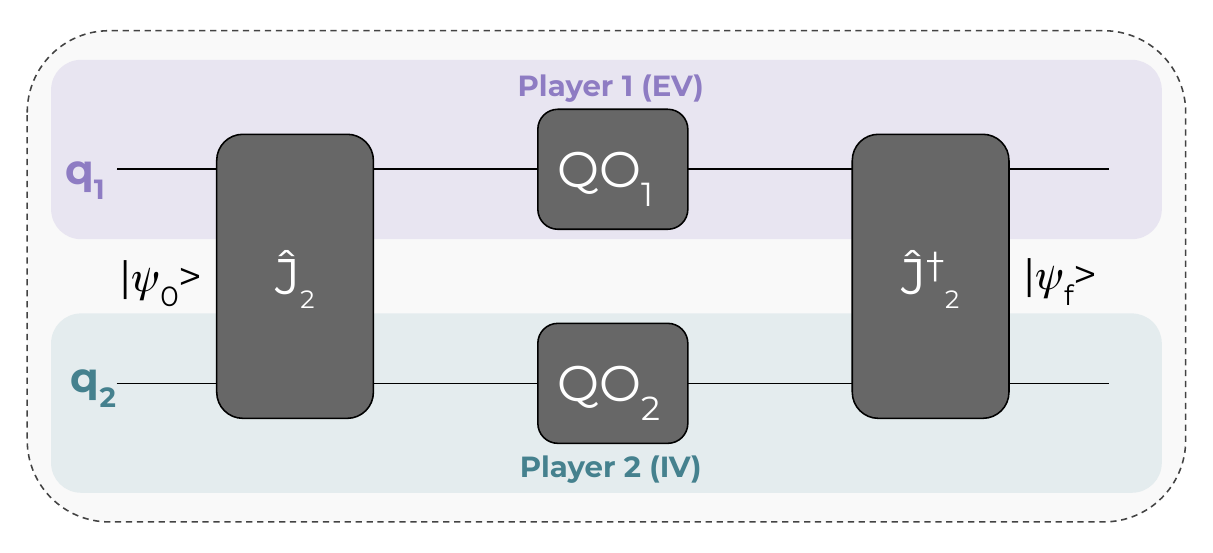}}\hfill
\subfigure[QC - 3 players and 2 strategies.]{\includegraphics[width=.33\textwidth]{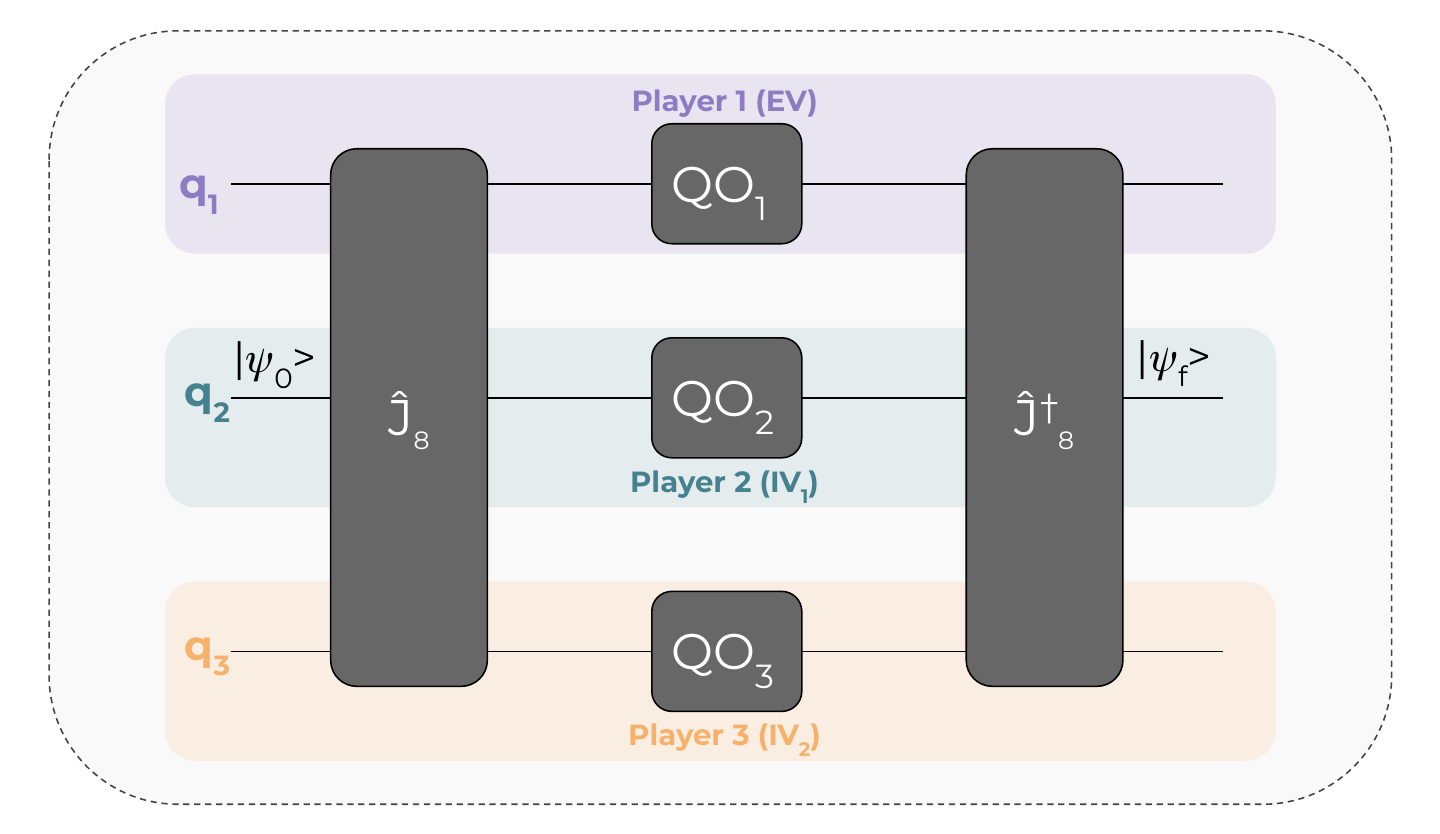}}\hfill 
\subfigure[QC - 2 players and 3 strategies.]{\includegraphics[width=.33\textwidth]{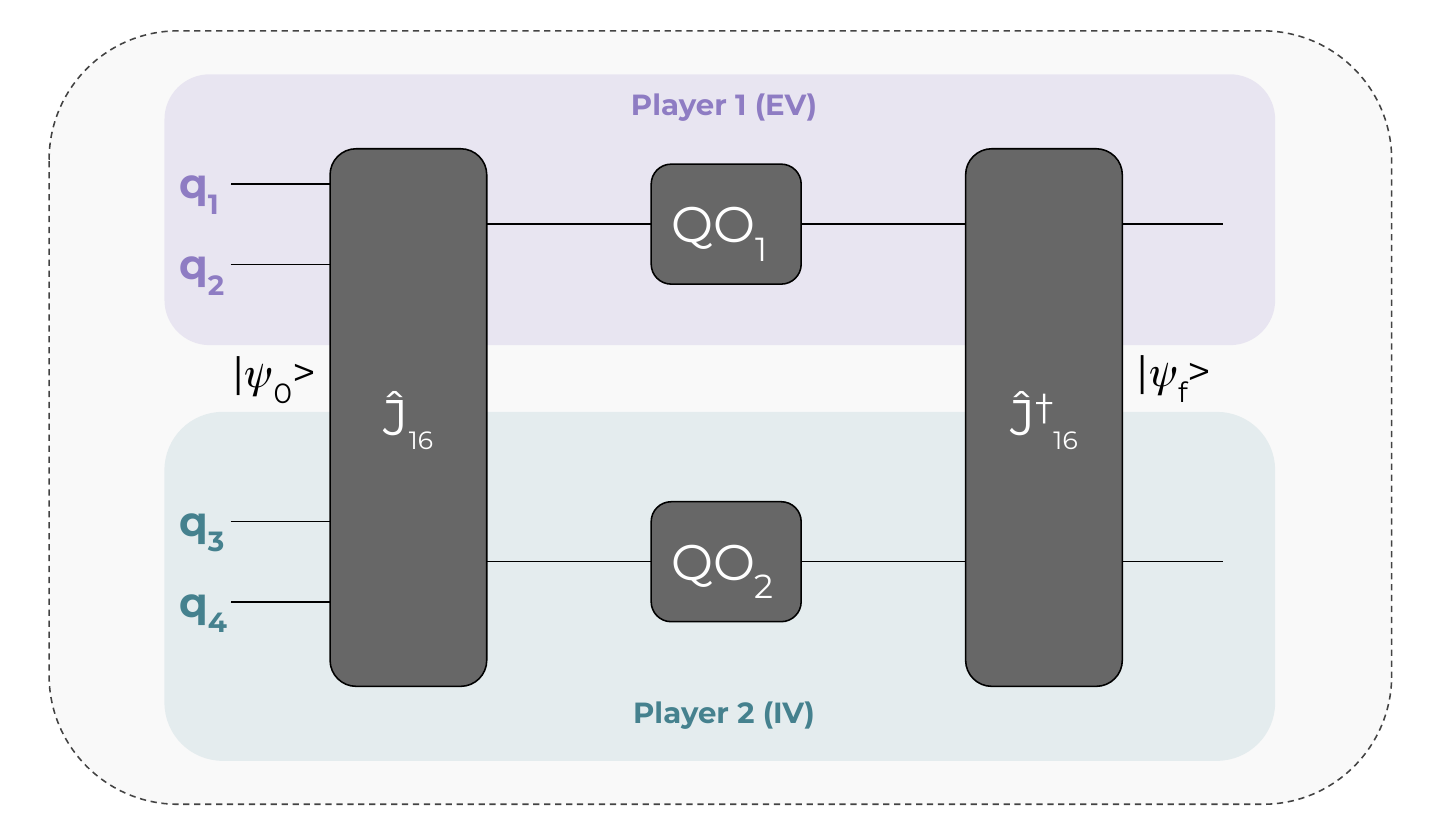}}\hfill
\caption{Quantum circuits used in the QGDM framework.}
\label{fig:quantumcircuits}
\end{figure*}

The first distinction between classical and quantum games lies in how strategies are represented. In classical games, a pure strategy can be encoded as a classical bit. In quantum games, however, it is represented as a qubit \(q_i\).
A qubit for player \(i\) is defined as:

\begin{equation} \label{eq:qubit}
    q_i= \alpha_i \big|0 \rangle + \beta_i \big| 1 \rangle \,,
\end{equation}
where \(\alpha_i, \beta_i \in \mathbb{C}\) are the complex probability amplitudes associated with the basis states. Here, \(\big|0 \rangle \) and \(\big|1 \rangle\) correspond to classical strategies \(a_i^1\) and \(a_i^2\) respectively. The coefficients \(\alpha_i\) and \(\beta_i\) represent the likelihood of adopting each strategy. Since they are probability amplitudes, they satisfy: 

\begin{equation} \label{eq:probability}
    \big|\alpha_i\big|^2 + \big|\beta_i\big|^2 = 1.
\end{equation}
The computational basis states are expressed as:

\begin{equation}
\begin{aligned}
    \big|0 \rangle &= 
    \begin{bmatrix}
        1 \\ 0
    \end{bmatrix} \,,
    &\quad\quad
    \big|1 \rangle &= 
    \begin{bmatrix}
        0 \\ 1
    \end{bmatrix}.
\end{aligned}
\end{equation}
As stated above, the quantum model outputs the probabilities of being in each state of the game. These probabilities are then used to compute the expected utility, which directly influences the maneuver (i.e., strategy) selected by the ego, since the objective is to maximize \( \mathbb{E}(u_i)\). 
As illustrated in Figure \ref{fig:quantumcircuits}, we propose three different Quantum Circuits (QC) depending on the scenario: (a) a model for two-player, two-strategy games, (b) a model for three-player, two-strategy games, and (c) a model for two-player, three-strategy games. For each QC, we build upon the Eisert-Wilkens-Lewenstein protocol \cite{eisert1999quantum}, adapting it to our context.

Each quantum circuit follows the same general structure. The variables \(\big| \psi_0 \rangle\) and \(\big| \psi_f \rangle\) represent, respectively, the initial and final quantum states of the system, which encode the likelihood of being in each possible state of the game. 
Also, each player \(i\) applies a quantum operator \(QO_i\) which corresponds to their chosen quantum strategy. We define two types of quantum operators:
\begin{itemize}
    \item \textbf{Unitary Operator \(\hat{U} (\theta_i)\)} (Eq. \ref{eq:unitary1parameters}), which controls the state of a qubit on the Bloch sphere through \(\theta_i \in [0,\pi]\). In fact, the variable \(\theta_i\) handles the degree of superposition between the player's strategies.

    \item \textbf{Quantum Gates \(QG_i\)}, which correspond to specific values of \(\theta_i\). These gates provide discrete strategic choices, offering simpler control of the qubits compared to a unitary operator. We defined \(QG_i = \{H,\sigma_x,\sigma_y,\sigma_z,I_2\}\), where:
    
    \begin{itemize}
        \item \(H\) is the Hadamard gate (Eq. \ref{eq:hadamard}),
        \item \(\sigma_x\), \(\sigma_y\), and \(\sigma_z\) are the Pauli gates (Eq. \ref{eq:Pauli}),
        \item \(I_2\) is the 2x2 identity matrix.
    \end{itemize}
\end{itemize}
Accordingly, since the qubits can be controlled either through the parameter \(\theta_i\) or through the quantum gates \(QG_i\), we distinguish two variants for each quantum circuit:

\begin{itemize}
    \item \textbf{QGDM - Unitary (QGDM-U)}: where players act through unitary operators \(U(\theta_i)\).
    \item \textbf{QGDM - Gate (QGDM-G)}: where players act through quantum gates \(QG_i\).
\end{itemize}

For the quantum circuit handling three players, we are inspired by the studies of \cite{dong2021superiority} and \cite{du2002entanglement} to account for an additional player.

Concerning the quantum circuit handling two players with three strategies per player, each player's set of strategies is encoded using two qubits, giving a total of \(N_q=4\) qubits. Since each player controls two qubits, the player's quantum operator \(QO_i\) is composed of two quantum operators: \(QO_i=U(\theta_i) \otimes U(\theta_i)\) for QGDM-U, and \(QO_i=QG_i \otimes QG_i\) for QGDM-G. This encoding allows each player to represent their three strategies.

A detailed description of each component of the quantum circuits and their role can be found in \cite{essalmiquantum}.

The main structural difference across the three proposed quantum circuits lies in the components \(\hat{J}(\gamma)\) (Eq. \ref{eq::J}) and \(\hat{J}^\dagger(\gamma)\), which represent, respectively, the entanglement and disentanglement operators. These components are parameterized by \(\gamma \in [0,\frac{\pi}{2}]\), which controls the degree of entanglement between qubits. Entanglement is a quantum aspect that captures the correlation between qubits: at maximum entanglement (\(\gamma=\frac{\pi}{2}\)), the qubits cannot be described independently, and the game becomes maximally entangled. In our case, the entanglement level controls how the strategy of player \(i\) influences, and is influenced by, the strategy of the opposing players \(-i\).
Mathematically, \(\hat{J}^\dagger\) is the Hermitian conjugate of \(\hat{J}\).

For a quantum game with \(N_p\) players, the final quantum state of the system is given by:

\begin{equation} \label{eq:psif}    
\big|\psi_f \rangle = \mathcal{\hat{J}}^{\dagger} \cdot 
\left( \bigotimes_{i=1}^{N_p} QO_i \right) \cdot \mathcal{\hat{J}} \cdot \big|\psi_0 \rangle\,,
\end{equation}
where \(\big|\psi_0 \rangle = \bigotimes_{i=1}^{N_q} q_i\) is the initial quantum state of the circuit, with \(N_q\) the total number of qubits. The dimensions of \(\big|\psi_f \rangle\) and \(\big|\psi_0 \rangle\) depend on the number of qubits, and are given by \(\big|\psi_f \rangle, \big|\psi_0 \rangle \in \mathbb{C}^{m}\), with \(m = 2^{N_q}\).
For example, in the quantum circuit handling 2 players and 2 strategies (modeled with 2 qubits), \(\big|\psi_f \rangle = [\psi_{f_{00}}, \psi_{f_{01}}, \psi_{f_{10}}, \psi_{f_{11}}]^T\), where \(|\psi_{f_{ij}}|^2\) represents the probability of being in state \(ij\), corresponding to player 1 playing strategy \(i\) and player 2 playing strategy \(j\).

\begin{equation} \label{eq::J}
    \hat{J}(\gamma) = \exp\Big(-i \frac{\gamma}{2} \bigotimes_{i=1}^{N_q} \sigma_x \Big)\,,
\end{equation}

\begin{equation} \label{eq:unitary1parameters}
    \hat{U} (\theta_i)=\begin{bmatrix}
            cos\frac{\theta}{2} & sin\frac{\theta}{2} \\
            -sin\frac{\theta}{2} & cos\frac{\theta}{2} 
            \end{bmatrix}\,,
\end{equation}
\begin{equation} \label{eq:hadamard}
    H = \frac{1}{\sqrt2} \begin{bmatrix}
        1 & 1 \\
        1 & -1
    \end{bmatrix}\,,
\end{equation}
\begin{equation} \label{eq:Pauli}
\begin{aligned}
    \sigma_x &= \begin{bmatrix}
        0 & 1 \\
        1 & 0
    \end{bmatrix}\,, \quad
    \sigma_y = \begin{bmatrix}
        0 & -j \\
        j & 0
    \end{bmatrix}\,, \quad
    \sigma_z = \begin{bmatrix}
        1 & 0 \\
        0 & -1
    \end{bmatrix}.
\end{aligned}
\end{equation}

\section{Results} \label{sec::results}

\begin{figure*}
    \centering
    \includegraphics[width=0.97\textwidth]{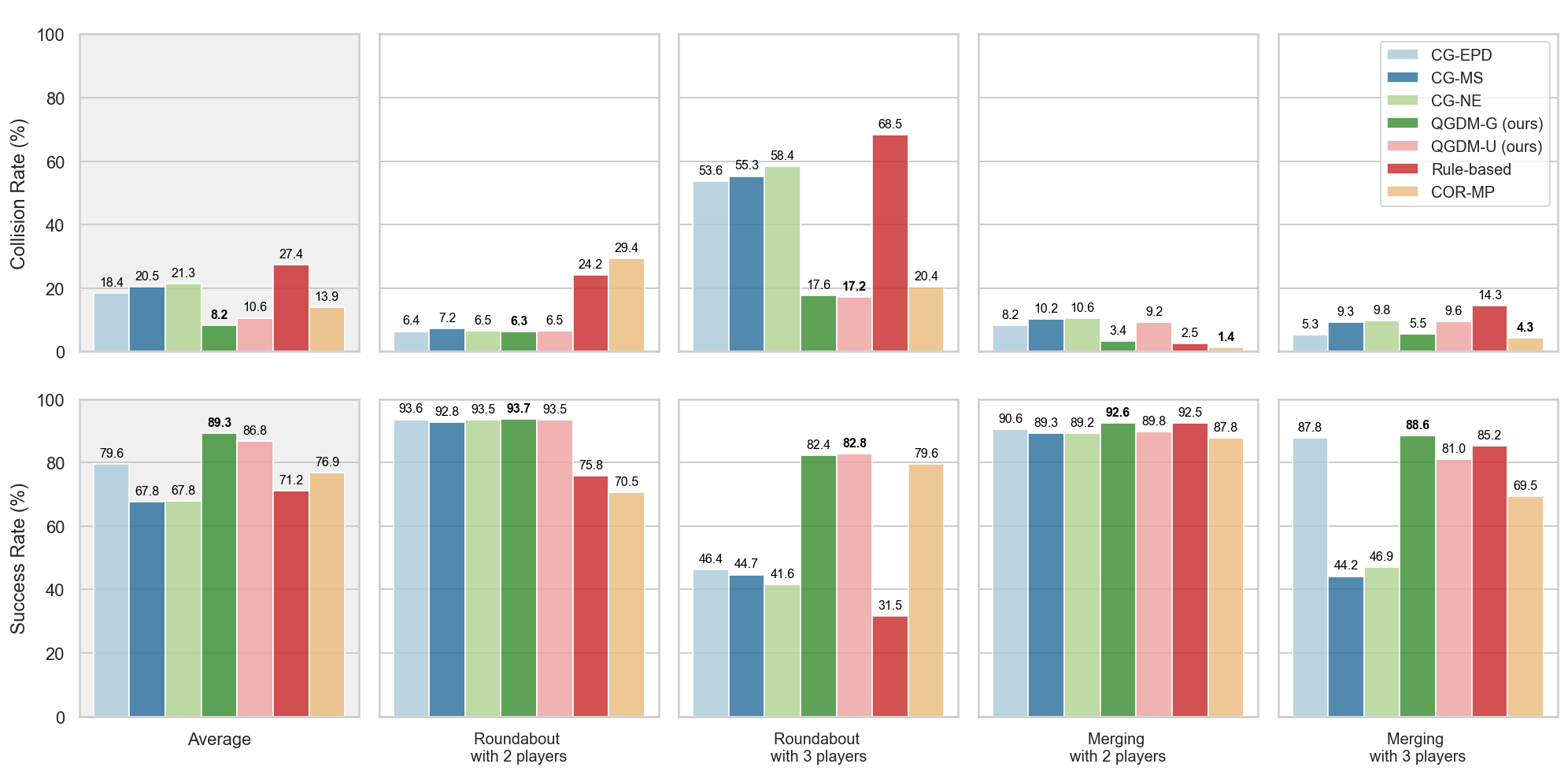}
    \caption{Quantitative results obtained for the merging and roundabout scenarios presented in this paper, compared against baseline methods. \textbf{CR} denotes the Collision Rate (percentage of episodes ending in a collision), and \textbf{SR} denotes the Success Rate (percentage of episodes in which the ego vehicle completes the episode without collision or becoming stuck). For the rule-based baseline, IDM \cite{IDM} was used in the roundabout scenarios and MOBIL \cite{MOBIL} in the merging scenarios.}
    \label{fig:results}
\end{figure*}

As described above, we evaluated our method on simulated scenarios using the highway-env simulator \cite{highway-env}. Our approach was tested in various scenarios, including roundabout and merging use cases with two and three players, as well as highway scenarios while considering 3 strategies per player. For each use case, we compared our method with existing approaches: the Intelligent Driver Model (IDM) \cite{IDM}, a well-known rule-based model that manages longitudinal actions; the Minimizing Overall Braking Induced by Lane changes (MOBIL) model, a rule-based model handling lateral actions; and COR-MP \cite{cor-mp}, a utility-based maneuver planner handling both longitudinal and lateral actions. We also include comparisons with classical game-theoretic methods: CG-MS (Classical Game - Mixed Strategy), where probabilities are computed through the mixed strategy technique, CG-EPD (Classical Game - Equal Probability Distribution), where probabilities are distributed equally among feasible actions, and CG-NE (Classical Game - Nash Equilibrium), where probabilities are distributed equally among existing Nash Equilibria. Concerning this model, if no N.E. exists, the probabilities follow an equal probability distribution across all feasible actions. 

\begin{figure}[ht]
\subfigure[Roundabout (2 Players).]{\includegraphics[width=.24\textwidth]{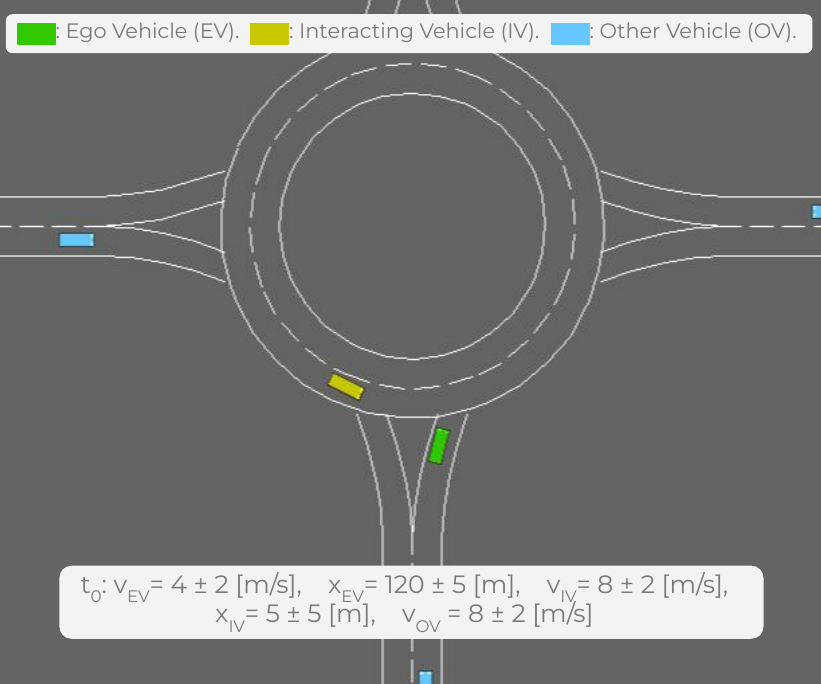}}\hfill
\subfigure[Roundabout (3 Players).]{\includegraphics[width=.24\textwidth]{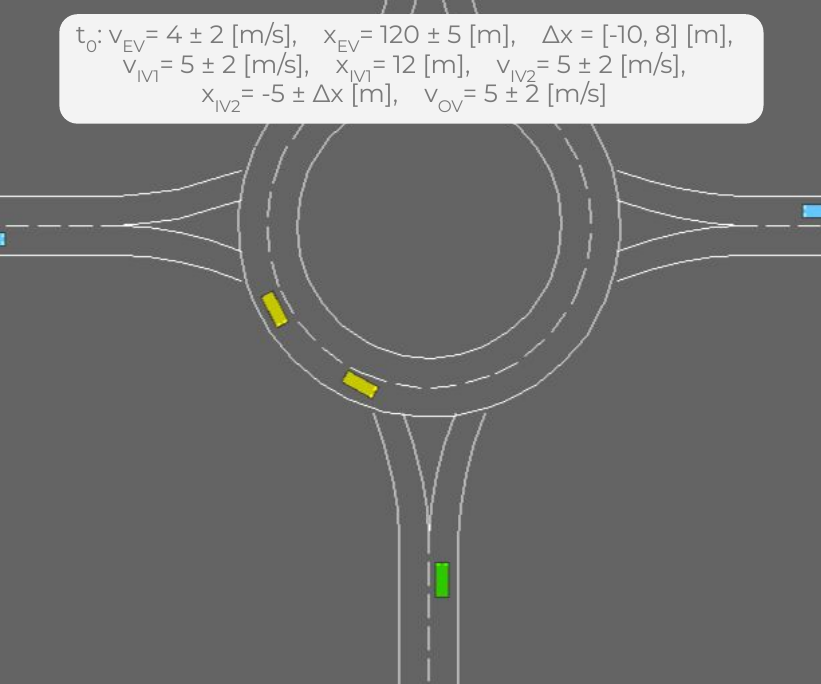}}\\
\subfigure[Merging (2 Players).]{\includegraphics[width=.5\textwidth]{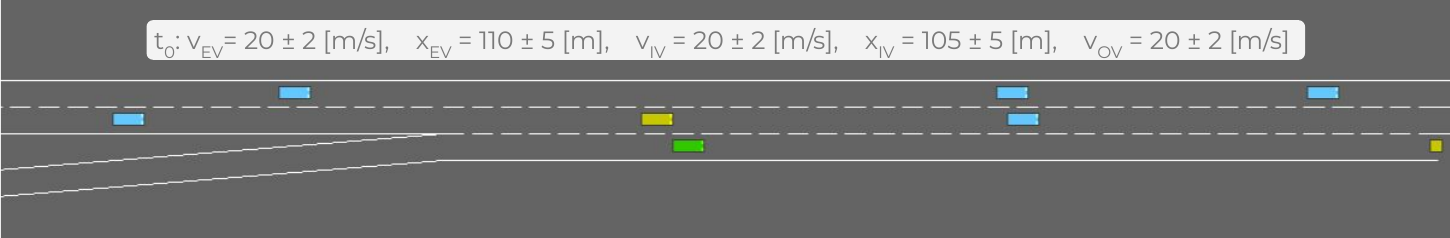}} \\
\subfigure[Merging (3 Players).]{\includegraphics[width=.5\textwidth]{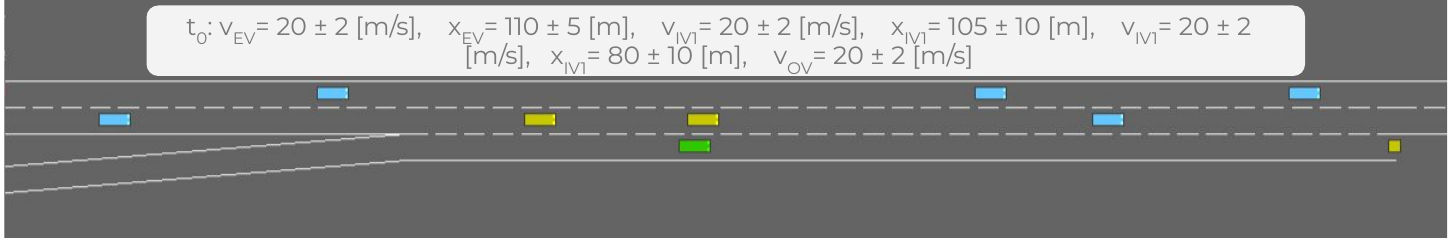}} \\
\subfigure[Highway.]{\includegraphics[width=.5\textwidth]{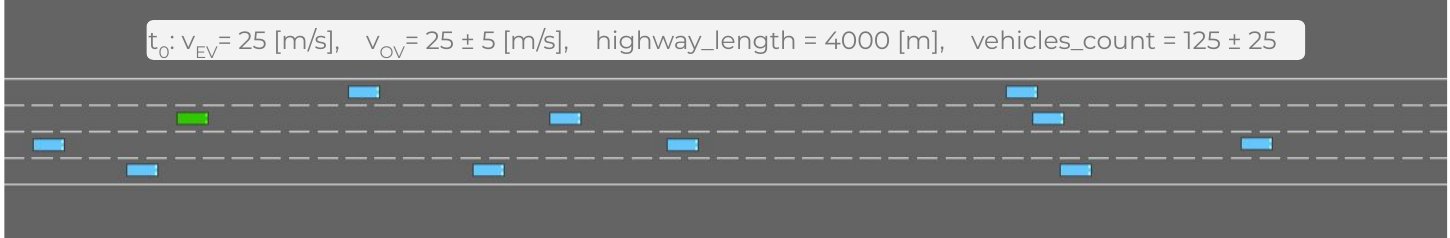}}
\caption{Different use cases presented in this paper with their respective initial setup parameters. 
}
\label{fig:scenarios}
\end{figure}

In the highway scenarios, the decisions of the Interacting Vehicle (IV) and of the Other Vehicles (OVs) are made using a combination of IDM \cite{IDM} and MOBIL \cite{MOBIL} models, enabling control of both longitudinal and lateral actions. The model parameters are set to represent a regular driver profile \cite{moghadam2021autonomous}. For the other scenarios, to introduce uncertainty in the behavior of IVs, their decisions are made according to an equal probability distribution over their respective action sets.
For each scenario, we run the simulation thousands of times, except for the highway scenario, which is run hundreds of times since it is a larger scenario. In every run, we vary the initial setup, including parameters such as ego speed, position, and positions of interacting vehicles. Figure \ref{fig:scenarios} shows each scenario and its varying initial parameters. Finally, regarding the quantum parameters of each model (quantum operators, entanglement level, and initial state), the best-performing configurations we found are summarized in Table \ref{tab:quantumparams}.

\begin{table}[hb!]
\centering
\begin{tabular}{l|l|l|l|l|l}
\toprule
\textbf{Model} & \textbf{Game Type} & $\mathbf{QO_{EV}}$ & $\mathbf{QO_{IV}}$ & $\mathbf{\gamma}$ & $\mathbf{\big| \psi_0 \big|^2}$\\
\midrule
QGDM-U & All & $\theta_{EV} = \frac{\pi}{2}$ & $\theta_{IV} = 0$ & 0 & EPD \\
\midrule
QGDM-Q & 2P / 2S & $I_2$ & $\sigma_y$ & $\frac{\pi}{2}$ & 
$[0,0,1,0]^T$ \\
QGDM-Q & Others & $H$ & $\sigma_x$ & $\frac{\pi}{3}$ & EPD \\
\bottomrule
\end{tabular}
\caption{Quantum parameters used for QGDM-U and QGDM-Q models. \textbf{2P / 2S} denotes games with two players and two strategies. \textbf{EPD} stands for Equal Probability Distribution. The initial quantum state \(\big| \psi_0 \big|^2 = [0,0,1,0]^T\) indicates that the game starts in the joint action profile where player 1 selects \(a_1^1\) and player 2 selects \(a_2^0\), corresponding, for example, to the ego decelerating and the interacting vehicle accelerating in the merging use case with two players.}
\label{tab:quantumparams}
\vspace{-0.5cm}
\end{table}

\begin{table*}[ht!]
    \centering
    \resizebox{\textwidth}{!}{
    \begin{tabular}{cccccccccc}
    \addlinespace
    \hline
    \textbf{Scenario} & \textbf{Method} & $\mathbf{n_{col}\downarrow}$ & \textbf{HD [m]} $\mathbf{\uparrow}$ & \(\mathbf{\bar{v}_{EV}}\) \textbf{[m/s]} $\mathbf{\uparrow}$ & \(\mathbf{\bar{a}_{EV}}\) \textbf{[m/s²]}& \(\mathbf{\bar{t}}\) \textbf{[s]} $\mathbf{\uparrow}$ & \(\mathbf{\rho_{CLL}}\) \textbf{[\%]} & \(\mathbf{\rho_{CLR}}\) \textbf{[\%]} & \(\mathbf{\rho_{KL}}\) \textbf{[\%]} \\
    \hline    
    \multirow{7}{*}{Highway} & COR-MP \cite{cor-mp} & 3 & \textbf{85.46} & 20.95 & -0.001 & 186.14 & 1.63 & 0.93 & 97.44 \\
    & IDMxMOBIL \cite{IDM,MOBIL} & \textbf{0} & 61.31 & 21.26 & -0.01 & \textbf{175.9} & 1.64 & 1.16 & 97.2 \\
    \cline{2-10}
    \addlinespace
    & CG-EPD & 1 & 27.5 & 21.88 & -0.002 & 209.3 & 3.5 & 2 & 94.5 \\
    & CG-MS & 1 & 26.95 & 21.69 & -0.003 & 210.8 & 2 & 2.9 & 95.1 \\
    & CG-NE & \textbf{0} & 27.6 & \textbf{22.02} & -0.008 & 219.1 & 3.9 & 2.7 & 93.4 \\
    \cline{2-10}
    \addlinespace
    & QGDM-G (ours)& \textbf{0} & 27.45 & 21.78 & \textbf{0} & 177.89 & 2.48 & 0.37 & 97.15 \\
    & QGDM-U (ours)& \textbf{0} & 25.98 & 21.6 & -0.004 & 249.6 & 2.5 & 1.2 & 96.3 \\
    \hline
    \end{tabular}
    }
    \caption{Quantitative results obtained for the highway scenario, compared against baseline methods. $\mathbf{n_{col}}$: number of collisions detected, \textbf{HD}: mean Headway Distance (between EV and vehicle ahead), \(\mathbf{\bar{v}_{EV}}\): mean ego speed, \(\mathbf{\bar{a}_{EV}}\): mean ego acceleration, \(\mathbf{\bar{t}}\): mean duration of an episode, \(\mathbf{\rho_{CLL}}\): proportion of left lane changes maneuvers, \(\mathbf{\rho_{CLR}}\): proportion of right lane changes maneuvers, and \(\mathbf{\rho_{KL}}\): proportion of keep lanes maneuvers.}
    \label{tab:quantitativeresults}
\end{table*}

\subsection{Merging}
We evaluated two different merging scenarios: one involving two players (subfigure (c) of Figure \ref{fig:scenarios}), where we use the quantum circuit depicted in subfigure (a) of Figure \ref{fig:quantumcircuits}, and another involving three players (subfigure (d) of Figure \ref{fig:scenarios}) with the quantum circuit shown in subfigure (b) of Figure \ref{fig:quantumcircuits}.
In both scenarios, the set of actions available to the ego vehicle is \(\mathcal{A}_{EV} = \{\text{Merge}, \text{Decelerate}\}\), while the interacting vehicle can choose \(\mathcal{A}_{IV} = \{\text{Accelerate}, \text{Decelerate}\}\).
The results are shown in Figure \ref{fig:results}, which highlights the collision rate observed across thousands of simulated episodes, as well as the success rate, defined as the proportion of time in which the ego successfully merges into the left lane. In some situations, the EV and IVs fail to synchronize, causing the ego to remain stuck in the merging lane. Among the tested methods, the utility-based one (COR-MP) yields the lowest collision rate but also the lowest success rate. This is because its strong emphasis on the safety aspect leads to overly conservative behavior, often leading to the ego being stuck. Conversely, the QGDM-G model achieves the highest success rate. 

\subsection{Roundabout}
Similar to the merging scenarios, we also considered two roundabout scenarios: one involving two players (subfigure (a) of Figure \ref{fig:scenarios}), using the quantum circuit shown in subfigure (a) of Figure \ref{fig:quantumcircuits}, and another involving three players (subfigure (b) of Figure \ref{fig:scenarios}) with the quantum circuit depicted in subfigure (b) of Figure \ref{fig:quantumcircuits}. However, the considered pure set of strategies is now: \(\mathcal{A}_{EV} = \mathcal{A}_{IV} = \{\text{Accelerate}, \text{Decelerate}\}\).
The results are presented in Figure \ref{fig:results}. In this case, the differences between methods are more clearly visible, particularly when moving from two to three players, where traditional methods fail to adapt effectively. Despite the occurrence of collisions justified by the irrationality of the IVs, the quantum models outperform the other approaches in terms of both collision and success rates for the roundabout scenarios.

Overall, as illustrated by the average plots (leftmost plots of Figure \ref{fig:results}), QGDM-G is the most scalable model, achieving the lowest average collision rate and the highest success rate across the roundabout and merging use cases. It demonstrates strong adaptability to uncertainty and potential irrational behavior in highly interactive situations, such as roundabouts and merging. These results further highlight the limitations of traditional, non-interaction-aware methods, which fail to adapt in highly interactive scenarios, as reflected in their higher collision rates and lower success rates. 

\subsection{Highway}
We also evaluated our approach in highway scenarios. Here, the EV can choose among \(\mathcal{A}_{EV}=\{\text{Change Lane Left}, \text{Change Lane Right}, \text{Idle}\}\), while the IV can select \(\mathcal{A}_{IV}=\{\text{Accelerate}, \text{Decelerate}, \text{Idle}\}\). The IV acting as the opponent player in the game is selected as the vehicle closest to the EV. The longitudinal behavior of the ego is controlled by the IDM model \cite{IDM}, with parameters set to represent a regular driver profile \cite{moghadam2021autonomous}. To account for more than two strategies per player, we employ the quantum circuit depicted in subfigure (c) of Figure \ref{fig:quantumcircuits}. 
For each episode, the initial lane of the ego vehicle is randomly selected among the 4 available ones.
The results from this simulation are shown in Table \ref{tab:quantitativeresults}. Here, we noticed several notable differences. First of all, the average time \(\mathbf{\bar{t}}\) to complete an episode is lowest for the IDMxMOBIL method, which is not surprising as this model is specifically designed to optimize efficiency on highways, and other road users are also modeled using IDMxMOBIL. COR-MP, the utility-based method, is the most conservative, with a mean headway distance \textbf{HD} of 85.46 meters. Overall, the quantum models did not produce any collisions in this scenario, demonstrating their applicability even in regular highway conditions where interactions are less intense compared to roundabout or merging situations. QGDM-U exhibits a higher \(\mathbf{\bar{t}}\) compared to QGDM-G and IDMxMOBIL due to performing additional lane changes (\(\rho_{CLL}\) and \(\rho_{CLR}\)), which occasionally cause the ego vehicle to become temporarily blocked behind another vehicle and unable to change lanes anymore.

\section{Conclusion} \label{sec::conclusion}
In this work, we introduced QGDM, a novel approach that combines game theory with quantum mechanics principles for interaction-aware decision-making in automated driving. QGDM runs in real time on a standard computer and does not require any quantum hardware. Our experiments show that traditional decision-making methods often lead to overly conservative behavior in highly interactive situations, whereas QGDM adapts effectively. It remains robust even when other road users exhibit uncertain or irrational behavior and performs comparably in scenarios where they act rationally, such as on highways. QGDM is easily scalable to more players and strategies by considering additional qubits, and its applicability is not limited to automated driving. It can be used for other robotic decision-making problems. 
Future work will extend the evaluation to dense urban environments and integrate QGDM with learning techniques to dynamically adapt quantum parameters to evolving traffic.



\bibliographystyle{IEEEtran}
\bibliography{bibliography}

@article{garrido2022review,
  title={{Review of decision-making and planning approaches in automated driving}},
  author={Garrido, Fernando and Resende, Paulo},
  journal={IEEE Access},
  year={2022},
  publisher={IEEE}
}

@inproceedings{cor-mp,
  title={{COR-MP: Conservation of Resources Model for Maneuver Planning}},
  author={Essalmi, Karim and Garrido, Fernando and Nashashibi, Fawzi},
  booktitle={2024 IEEE 20th International Conference on Intelligent Computer Communication and Processing (ICCP)},
  pages={1--8},
  year={2024},
  organization={IEEE}
}

@INPROCEEDINGS{cor-mcts,
  author={Essalmi, Karim and Garrido, Fernando and Nashashibi, Fawzi},
  booktitle={2025 IEEE Intelligent Vehicles Symposium (IV)}, 
  title={{An Extended Horizon Tactical Decision-Making for Automated Driving Based on Monte Carlo Tree Search}}, 
  year={2025},
  volume={},
  number={},
  pages={1127-1132},
  doi={10.1109/IV64158.2025.11097338}}

@article{MOBIL,
  title={{General lane-changing model MOBIL for car-following models}},
  author={Kesting, Arne and Treiber, Martin and Helbing, Dirk},
  journal={Transportation Research Record},
  volume={1999},
  year={2007}
}

@article{IDM,
  title={{Congested traffic states in empirical observations and microscopic simulations}},
  author={Treiber, Martin and Hennecke, Ansgar and Helbing, Dirk},
  journal={Physical review E},
  year={2000}
}

@misc{highway-env,
  author = {Leurent, Edouard},
  title = {{An Environment for Autonomous Driving Decision-Making}},
  year = {2018},
  publisher = {GitHub},
  journal = {GitHub repository},
  howpublished = {\url{https://github.com/eleurent/highway-env}},
}

@article{jackson2011brief,
  title={{A brief introduction to the basics of game theory}},
  author={Jackson, Matthew O},
  journal={Available at SSRN 1968579},
  year={2011}
}

@article{hu2025survey,
  title={{A survey of decision-making and planning methods for self-driving vehicles}},
  author={Hu, Jun and Wang, Yuefeng and Cheng, Shuai and Xu, Jinghan and Wang, Ningjia and Fu, Bingjie and Ning, Zuotao and Li, Jingyao and Chen, Hualin and Feng, Chaolu and others},
  journal={Frontiers in Neurorobotics},
  year={2025},
  publisher={Frontiers Media SA}
}

@article{makahleh2024assessing,
  title={{Assessing the role of autonomous vehicles in urban areas: A systematic review of literature}},
  author={Makahleh, Hisham Y and Ferranti, Emma Jayne Sakamoto and Dissanayake, Dilum},
  journal={Future Transportation},
  volume={4},
  number={2},
  pages={321--348},
  year={2024},
  publisher={MDPI}
}

@article{tyagi2021autonomous,
  title={{Autonomous Intelligent Vehicles (AIV): Research statements, open issues, challenges and road for future}},
  author={Tyagi, Amit Kumar and Aswathy, SU},
  journal={International Journal of Intelligent Networks},
  volume={2},
  pages={83--102},
  year={2021},
  publisher={Elsevier}
}

@article{malik2022autonomous,
  title={{How do autonomous vehicles decide?}},
  author={Malik, Sumbal and Khan, Manzoor Ahmed and El-Sayed, Hesham and Khan, Jalal and Ullah, Obaid},
  journal={Sensors},
  volume={23},
  number={1},
  pages={317},
  year={2022},
  publisher={MDPI}
}

@article{pan2025tgld,
  title={{TGLD: A Trust-Aware Game-Theoretic Lane-Changing Decision Framework for Automated Vehicles in Heterogeneous Traffic}},
  author={Pan, Jie and Wang, Tianyi and Wang, Yangyang and Jiao, Junfeng and Claudel, Christian},
  journal={arXiv preprint arXiv:2507.10075},
  year={2025}
}

@inproceedings{trautman2010unfreezing,
  title={{Unfreezing the robot: Navigation in dense, interacting crowds}},
  author={Trautman, Peter and Krause, Andreas},
  booktitle={2010 IEEE/RSJ International Conference on Intelligent Robots and Systems},
  pages={797--803},
  year={2010},
  organization={IEEE}
}

@inproceedings{fisac2019hierarchical,
  title={{Hierarchical game-theoretic planning for autonomous vehicles}},
  author={Fisac, Jaime F and Bronstein, Eli and Stefansson, Elis and Sadigh, Dorsa and Sastry, S Shankar and Dragan, Anca D},
  booktitle={2019 International conference on robotics and automation (ICRA)},
  pages={9590--9596},
  year={2019},
  organization={IEEE}
}

@incollection{von2007theory,
  title={{Theory of games and economic behavior: 60th anniversary commemorative edition}},
  author={Von Neumann, John and Morgenstern, Oskar},
  booktitle={Theory of games and economic behavior},
  year={2007},
  publisher={Princeton university press}
}

@article{nosheen2025game,
  title={{GAME THEORY: FOUNDATIONS, APPLICATIONS, AND FUTURE PERSPECTIVES}},
  author={Nosheen, Misbah and Sami, Fariha and Rasheed, Iqra},
  journal={International Journal of Social Sciences Bulletin},
  year={2025}
}

@article{zhang2021subjective,
  title={{A subjective model of human decision making based on quantum decision theory}},
  author={Zhang, Chenda and Kjellstr{\"o}m, Hedvig},
  journal={arXiv preprint arXiv:2101.05851},
  year={2021}
}

@article{eisert1999quantum,
  title={{Quantum games and quantum strategies}},
  author={Eisert, Jens and Wilkens, Martin and Lewenstein, Maciej},
  journal={Physical Review Letters},
  year={1999},
  publisher={APS}
}

@article{wang2022advantages,
  title={{ADVANTAGES AND APPLICATIONS OF QUANTUM GAME THEORY}},
  author={WANG, HAOSHU},
  year={2022}
}

@misc{wiki_quantum_game_theory,
  author       = {{Wikipedia contributors}},
  title        = {{Quantum game theory}},
  url          = {https://en.wikipedia.org/wiki/Quantum\_game\_theory},
}

@article{pricequantum,
  title={{Quantum Games and Game Strategy}},
  author={Price, Elizabeth}
}

@article{lukasik2018quantum,
  title={{Quantum models of cognition and decision}},
  author={{\L}ukasik, Andrzej},
  journal={International journal of parallel, emergent and distributed systems},
  year={2018}
}

@article{morra2023quantum,
  title={{Quantum game theory}},
  author={Morra, Giuseppe Tancredi},
  year={2023}
}

@misc{talavera2021autonomous,
  title={{Autonomous vehicles technological trends}},
  author={Talavera, Edgar and D{\'\i}az-{\'A}lvarez, Alberto and Naranjo, Jos{\'e} Eugenio and Olaverri-Monreal, Cristina},
  journal={Electronics},
  volume={10},
  number={10},
  pages={1207},
  year={2021},
  publisher={MDPI}
}

@inproceedings{ghoul2023interpretable,
  title={{Interpretable goal-based model for vehicle trajectory prediction in interactive scenarios}},
  author={Ghoul, Amina and Yahiaoui, Itheri and Verroust-Blondet, Anne and Nashashibi, Fawzi},
  booktitle={2023 IEEE Intelligent Vehicles Symposium}
}

@inproceedings{jaritz2018end,
  title={{End-to-end race driving with deep reinforcement learning}},
  author={Jaritz, Maximilian and De Charette, Raoul and Toromanoff, Marin and Perot, Etienne and Nashashibi, Fawzi},
  booktitle={2018 IEEE international conference on robotics and automation (ICRA)}
}

@article{pang2024large,
  title={{Large language model guided deep reinforcement learning for decision making in autonomous driving}},
  author={Pang, Hao and Wang, Zhenpo and Li, Guoqiang},
  journal={arXiv preprint arXiv:2412.18511},
  year={2024}
}

@inproceedings{aksjonov2021rule,
  title={{Rule-based decision-making system for autonomous vehicles at intersections with mixed traffic environment}},
  author={Aksjonov, Andrei and Kyrki, Ville},
  booktitle={2021 IEEE International Intelligent Transportation Systems Conference (ITSC)},
  pages={660--666},
  year={2021},
  organization={IEEE}
}

@inproceedings{de2020ethical,
  title={{Ethical decision making for autonomous vehicles}},
  author={De Moura, Nelson and Chatila, Raja and Evans, Katherine and Chauvier, St{\'e}phane and Dogan, Ebru},
  booktitle={2020 IEEE intelligent vehicles symposium (iv)},
  pages={2006--2013},
  year={2020},
  organization={IEEE}
}

@inproceedings{hubmann2019pomdp,
  title={{A POMDP maneuver planner for occlusions in urban scenarios}},
  author={Hubmann, Constantin and Quetschlich, Nils and Schulz, Jens and Bernhard, Julian and Althoff, Daniel and Stiller, Christoph},
  booktitle={2019 IEEE Intelligent Vehicles Symposium (IV)}
}

@inproceedings{klimenko2014tapir,
  title={{Tapir: A software toolkit for approximating and adapting pomdp solutions online}},
  author={Klimenko, Dimitri and Song, Joshua and Kurniawati, Hanna},
  booktitle={Proceedings of the Australasian Conference on Robotics and Automation, Melbourne, Australia},
  volume={24},
  year={2014}
}

@article{che2024game,
  title={{Game theory: Concepts, applications, and insights from operations research}},
  author={Che, Chang and Tian, Junchi},
  journal={Journal of Computer Technology and Applied Mathematics},
  volume={1},
  number={4},
  pages={53--59},
  year={2024}
}

@article{schoemaker1982expected,
  title={{The expected utility model: Its variants, purposes, evidence and limitations}},
  author={Schoemaker, Paul JH},
  journal={Journal of economic literature},
  year={1982}
}

@article{dong2021superiority,
  title={{The Superiority of Quantum Strategy in 3-Player Prisoner’s Dilemma}},
  author={Dong, Zhiyuan and Wu, Ai-Guo},
  journal={Mathematics},
  year={2021},
  publisher={MDPI}
}

@article{du2002entanglement,
  title={{Entanglement enhanced multiplayer quantum games}},
  author={Du, Jiangfeng and Li, Hui and Xu, Xiaodong and Zhou, Xianyi and Han, Rongdian},
  journal={Physics Letters A},
  year={2002},
  publisher={Elsevier}
}

@article{arbabi2023decision,
  title={{Decision making for autonomous driving in interactive merge scenarios via learning-based prediction}},
  author={Arbabi, Salar and Tavernini, Davide and Fallah, Saber and Bowden, Richard},
  journal={arXiv preprint arXiv:2303.16821},
  year={2023}
}

@article{heshami2024towards,
  title={{Towards Self-Organizing connected and autonomous Vehicles: A coalitional game theory approach for cooperative Lane-Changing decisions}},
  author={Heshami, Seiran and Kattan, Lina},
  journal={Transportation Research Part C: Emerging Technologies},
  volume={166},
  pages={104789},
  year={2024},
  publisher={Elsevier}
}

@inproceedings{chen2021midas,
  title={{MIDAS: Multi-agent interaction-aware decision-making with adaptive strategies for urban autonomous navigation}},
  author={Chen, Xiaoyi and Chaudhari, Pratik},
  booktitle={2021 IEEE International Conference on Robotics and Automation (ICRA)},
  pages={7980--7986},
  year={2021},
  organization={IEEE}
}

@article{nan2023interaction,
  title={{Interaction-aware planning with deep inverse reinforcement learning for human-like autonomous driving in merge scenarios}},
  author={Nan, Jiangfeng and Deng, Weiwen and Zhang, Ruzheng and Wang, Ying and Zhao, Rui and Ding, Juan},
  journal={IEEE Transactions on Intelligent Vehicles},
  volume={9},
  number={1},
  pages={2714--2726},
  year={2023},
  publisher={IEEE}
}

@article{li2023autonomous,
  title={{Autonomous Interactive Driving Decision Based on Global Sorting and Local Gaming}},
  author={Li, Daofei and Zhang, Jiajie and others},
  journal={Authorea},
  year={2023},
}

@inproceedings{garzon2019game,
  title={{Game theoretic decision making for autonomous vehicles’ merge manoeuvre in high traffic scenarios}},
  author={Garz{\'o}n, Mario and Spalanzani, Anne},
  booktitle={2019 IEEE Intelligent Transportation Systems Conference (ITSC)}
}

@inproceedings{moghadam2021autonomous,
  title={{An autonomous driving framework for long-term decision-making and short-term trajectory planning on frenet space}},
  author={Moghadam, Majid and Elkaim, Gabriel Hugh},
  booktitle={2021 IEEE 17th international conference on automation science and engineering (CASE)},
  year={2021},
  organization={IEEE}
}

@article{khan2025quantum,
  title={{Quantum Advantage in Trading: A Game-Theoretic Approach}},
  author={Khan, Faisal Shah and Linke, Norbert M and Than, Anton Trong and Baron, Dror},
  journal={Quantum Economics and Finance},
  volume={2},
  number={1},
  pages={40--51},
  year={2025},
  publisher={SAGE Publications Sage CA: Los Angeles, CA}
}

@article{meyer1999quantum,
  title={{Quantum strategies}},
  author={Meyer, David A},
  journal={Physical Review Letters},
  year={1999},
  publisher={APS}
}

@article{song2022quantum,
  title={{Quantum decision making in automatic driving}},
  author={Song, Qingyuan and Fu, Weiping and Wang, Wen and Sun, Yuan and Wang, Denggui and Zhou, Jincao},
  journal={Scientific reports},
  year={2022},
  publisher={Nature Publishing Group UK London}
}

@article{pothos2009quantum,
  title={{A quantum probability explanation for violations of ‘rational’decision theory}},
  author={Pothos, Emmanuel M and Busemeyer, Jerome R},
  journal={Proceedings of the Royal Society B: Biological Sciences},
  year={2009},
  publisher={The Royal Society London}
}

@article{song2022research,
  title={{Research on quantum cognition in autonomous driving}},
  author={Song, Qingyuan and Wang, Wen and Fu, Weiping and Sun, Yuan and Wang, Denggui and Gao, Zhiqiang},
  journal={Scientific reports},
  year={2022},
  publisher={Nature Publishing Group UK London}
}

@inproceedings{essalmiquantum,
  title={{Quantum game models for interaction-aware decision-making in automated driving}},
  author={Essalmi, Karim and Garrido, Fernando and Nashashibi, Fawzi},
  booktitle={2025 IEEE International Conference on Advanced Robotics (ICAR)},
  pages={155--162},
  year={2025},
  organization={IEEE}
}

\end{document}